\pdfoutput=1

\documentclass[11pt]{article}

\usepackage[]{ACL2023}

\usepackage{times}
\usepackage{latexsym}

\usepackage[T1]{fontenc}

\usepackage[utf8]{inputenc}

\usepackage{microtype}

\usepackage{inconsolata}
\usepackage{microtype}
\usepackage{booktabs}
\usepackage{multirow}
\usepackage{url}
\usepackage{tabularx}

\usepackage{amsmath}
\usepackage{amssymb}
\usepackage{mathtools}
\usepackage{amsthm}
\usepackage{mathabx}
\usepackage{soul}
\usepackage{url}
\usepackage{graphicx}
\usepackage{booktabs}
\usepackage{algorithm}
\usepackage{array}
\usepackage{comment}
\usepackage{amsfonts}
\usepackage{multirow}
\usepackage{dblfloatfix}
\usepackage{multicol}
\usepackage{xcolor}
\usepackage{upgreek}
\usepackage{siunitx}
\usepackage{natbib}
\usepackage{xspace}
\usepackage{adjustbox}
\usepackage{wrapfig}
\usepackage{bm}
\usepackage{mathrsfs}
\usepackage{amsfonts}
\usepackage{color}
\usepackage{colortbl}
\usepackage{CJKutf8}
\usepackage{booktabs}
\usepackage{array} 
\usepackage{mwe}
\usepackage{titlesec}
\usepackage{enumitem}
\usepackage{microtype}
\usepackage{fancypar}
\usepackage{extarrows}
\usepackage{array}
\usepackage{nicefrac}       
\usepackage{supertabular,enumitem}
\usepackage{hyperref}

\usepackage{epsfig}
\usepackage{changepage}
\usepackage{enumerate}
\usepackage{amsfonts,lscape,color,url}
\usepackage{tabularx}
\usepackage{adjustbox}
\usepackage{overpic}
\usepackage{tablefootnote}

\newcommand{\ModelName}{Mixture-of-Domain-Adapters}
\newcommand{\ModelNameShort}{MixDA}
\newcommand{\ModuleName}{domain-adapter}

\title{{\ModelName}: Decoupling and Injecting Domain Knowledge to Pre-trained Language Models' Memories}

\author{Shizhe Diao$^{\heartsuit*}$, ~ Tianyang Xu$^{\spadesuit*}$, ~ Ruijia Xu$^{\heartsuit}$, ~ Jiawei Wang$^{\clubsuit}$, ~ \bf Tong Zhang$^{\heartsuit}$\\
  $^{\heartsuit}$The Hong Kong University of Science and Technology\\
  \texttt{\{sdiaoaa, rxuaq, tongzhang\}@ust.hk}\\
  $^{\spadesuit}$Wuhan University\\
    \texttt{tianyangxu@whu.edu.cn}\\
  $^{\clubsuit}$Shanghai Jiao Tong University \\
  \texttt{wjw\_sjt@sjtu.edu.cn} \\
  \\
}

\begin{document}
\maketitle
\def\thefootnote{*}\footnotetext{Equal Contribution.}
\def\thefootnote{\arabic{footnote}}

\begin{abstract}
Pre-trained language models (PLMs) demonstrate excellent abilities to understand texts in the generic domain while struggling in a specific domain.
Although continued pre-training on a large domain-specific corpus is effective, it is costly to tune all the parameters on the domain.
In this paper, we investigate whether we can adapt PLMs both effectively and efficiently by only tuning a few parameters. 
Specifically, we decouple the feed-forward networks (FFNs) of the Transformer architecture into two parts: the original pre-trained FFNs to maintain the old-domain knowledge and our novel domain-specific adapters to inject domain-specific knowledge in parallel.
Then we adopt a mixture-of-adapters gate to fuse the knowledge from different domain adapters dynamically. 
Our proposed {\ModelName} (\textbf{\ModelNameShort}) employs a two-stage adapter-tuning strategy that leverages both unlabeled data and labeled data to help the domain adaptation: $i$) domain-specific adapter on unlabeled data;
followed by $ii$) the task-specific adapter on labeled data.   
{\ModelNameShort} can be seamlessly plugged into the pretraining-finetuning paradigm and our experiments demonstrate that {\ModelNameShort} achieves superior performance on in-domain tasks~(GLUE), out-of-domain tasks~(ChemProt, RCT, IMDB, Amazon), and knowledge-intensive tasks~(KILT).
Further analyses demonstrate the reliability, scalability, and efficiency of our method.\footnote{The code is available at \url{https://github.com/Amano-Aki/Mixture-of-Domain-Adapters}.}
\end{abstract}

\section{Introduction}
Pre-trained language models (PLMs) have achieved a multitude of successful applications in natural language understanding~\citep{devlin2018bert, liu2019roberta, he2021deberta} and generation~\citep{lewis2019bart, zhang2019dialogpt, yang2020styledgpt, NEURIPS2020_1457c0d6}.
The predominant methodology for domain adaptation is fine-tuning on labeled domain-specific data or continued pre-training~\citep{gururangan2020don} on unlabeled domain-specific data.
Although effective, both fine-tuning and continued pre-training methods require tuning all the parameters of a PLM, raising high costs beyond many institutions' reach.
To mitigate this, multiple parameter-efficient fine-tuning (PEFT) methods are proposed, including prompt-based tuning~\citep{gao2020making, liu2021gpt, schick2021s, Li2021Prefix, liu2021pre}, and adapter-based tuning~\citep{Houlsby2019Feb, pfeiffer-etal-2020-mad, Hu2021LoRA}. 
However, they are more concerned about task adaptation and it is still unclear how to regularly, and inexpensively inject domain knowledge into PLMs for different domain-specific tasks.
Moreover, directly tuning PLMs on a domain-specific corpus with PEFT methods will lead to the catastrophic forgetting problem~\citep{yogatama2019learning, gururangan2020don}. 
These limitations highlight an important research question: \emph{how to adapt PLMs with the new domain knowledge while keeping the old-domain knowledge unmodified?}

Inspired by the recent studies~\citep{geva2021transformer, decao2021editing, Meng2022Rome} that found knowledge is stored in feed-forward networks (FFNs), we decouple the FFNs into two parts: the original pre-trained FFNs to maintain the old-domain knowledge and our novel domain-specific adapters to inject domain-specific knowledge in parallel.
Specifically, we propose {\ModelName} ({\ModelNameShort}), a mixture of several domain adapters to inject domain-specific knowledge without affecting the old-domain knowledge.
Our model has two stages: $(i)$ domain-specific tuning multiple knowledge adapters on unlabeled data and then $(ii)$ task-specific tuning adapters on labeled data.
In the first stage, we train several domain adapters on both domain-specific corpus and pre-training corpus simultaneously while keeping the original feed-forward networks unchanged.
In the second stage, we train a mixture-of-adapters gate to dynamically select the desired knowledge adapter and a task-specific adapter for task adaptation.

We conduct experiments on a broad range of tasks, including 4 out-of-domain datasets, 9 in-domain datasets, and 2 knowledge-intensive datasets.
Our experimental results demonstrate the effectiveness of {\ModelNameShort} on 15 datasets, spanning biomedical, computer science publications, news, and reviews.
Further analysis displays three key properties of our proposed approach:
$(i)$ \textbf{Reliability}:
it shows superior performance on both in-domain and out-of-domain tasks.
$(ii)$ \textbf{Scalability}: it scales well to the increasing number of domains.
$(iii)$ \textbf{Efficiency}: it adds only a small number of parameters per domain.
We claim that these properties are helpful for language models as a service, where a 
frozen PLM is served, and multiple adapters are inserted to support different customized services.

\section{Related Work}\label{sec:related}
In this section, we will review four research lines related to injecting domain knowledge into pre-trained language models: knowledge injection, domain adaptation, parameter-efficient fine-tuning, and mixture-of-adapters.

\subsection{Knowledge Injection}
Knowledge can be injected into PLMs by pre-training or fine-tuning, each corresponding to a separate research direction.
During pre-training, the knowledge carried by knowledge graphs~\citep{zhang2019ernie, he2020bert}, entities~\citep{sun2019ernie, Xiong2020Pretrained}, n-grams~\citep{diao2020zen}, knowledge embedding~\citep{wang2021kepler}, synonym and hyponym-hypernym relations in 
WordNet~\citep{lauscher2019informing}, word-supersense knowledge~\citep{levine2020sensebert}, and knowledge bases~\citep{peters2019knowledge} can be injected into PLMs by feeding knowledge inputs and designing new objectives.
However, pre-training-based methods are costly, making the application to huge PLMs (e.g., models with 175 Billion parameters) impossible.
Fine-tuning-based methods only require an additional fine-tuning process.
Some studies inject extra information into the input sentences, like knowledge triples from knowledge graphs~\citep{liu2020k} and knowledge context~\citep{faldu2021ki}, while other studies explored specific model and training designs, like knowledge adapter networks~\citep{wang2021k}, graph convolutional networks and LSTMs~\citep{lin2019kagnet}, and meta-learning~\citep{Sinitsin2020Editable}.
\citet{zhu2020modifying} formulated knowledge injection as a constrained optimization problem by adding a constraint on the loss on the unmodified facts.
Recent studies~\citep{geva2021transformer, decao2021editing, Meng2022Rome} reveal that knowledge is stored in the feed-forward networks in PLMs.
Inspired by these studies, we propose a new efficient tuning method to inject domain knowledge into feed-forward networks with minimal costs.

\subsection{Domain Adaptation}
Previous studies have observed that language models suffer from a significant performance drop during the domain shift~\citep{beltagy2019scibert, alsentzer2019publicly, huang2019clinicalbert, lee2020biobert, ke2022adapting}.
Effective strategies that can bridge the domain gap are introduced.
Pre-training language models from scratch is effective but costly, like SciBERT~\citep{beltagy2019scibert}, BioBERT~\citep{lee2020biobert}, and ClinicalBERT~\citep{alsentzer2019publicly}.
Recent studies explored continued pre-training~\citep{gururangan2020don} and adapter networks~\citep{diao2021taming} to save time by training on unlabeled downstream task data.
In this paper, we introduce plug-in domain adaptors for domain adaptation, which are effective and mitigate catastrophic forgetting issues because of the explicit learning strategy and efficient model architecture.

\subsection{Parameter-Efficient Fine-tuning}
Another relevant research direction is parameter-efficient fine-tuning (PEFT), which only fine-tunes a small number of parameters.
Existing works solve this problem from two perspectives: prompt-based tuning~\citep{gao2020making, liu2021gpt, schick2021s, Li2021Prefix, liu2021pre}, and adapter-based tuning~\citep{Houlsby2019Feb, pfeiffer-etal-2020-mad, Hu2021LoRA}. 
Several works in adapter-based tuning are closely related to ours.
AdapterFusion~\citep{pfeiffer2021adapterfusion} aims to combine multiple task adapters but does not offer specific architecture or training strategies to learn external knowledge.
DEMix~\citep{gururangan2022demix} and MixDA both train adapters that specialize in domains and use mechanisms to route different adapters, but differ in routing methods, base models, and training strategies.
K-Adapter~\citep{wang2021k} is restricted by its training on T-REx triples and lacks the flexibility to train on unstructured knowledge.
Similar to MixDA, CPT~\citep{ke-etal-2022-continual} integrates domain knowledge into LMs, but it employs a different approach. While MixDA uses domain adapters to substitute FFN layers and task adapters to perform end tasks, CPT adds CL-Plugins that learn domain knowledge.
Recent work by~\citet{he2021towards} presents a unified framework that establishes connections across different PEFT methods.
Our work can leverage any PEFT method and complement them.

\subsection{Mixture-of-Experts}
Mixture-of-Experts (MoE)~\citep{shazeer2017outrageously} is introduced with several expert networks, gating networks, and load-balancing techniques.
The following studies improve MoE on initialization and training schemes~\citep{fedus2022switch}, routing mechanisms~\citep{zuo2021taming, yang2021m6}, and load-balancing issues~\citep{lewis2021base, roller2021hash}.
AdaMix~\citep{wang2022adamix} proposed a mixture of adapters to improve the downstream task performance.
Instead of mixing different designs of adapters, our domain adapter is a feed-forward network specifically designed for domain knowledge.

\section{Approach}
\begin{figure*}[t]
    \centering
    \includegraphics[width=\textwidth]{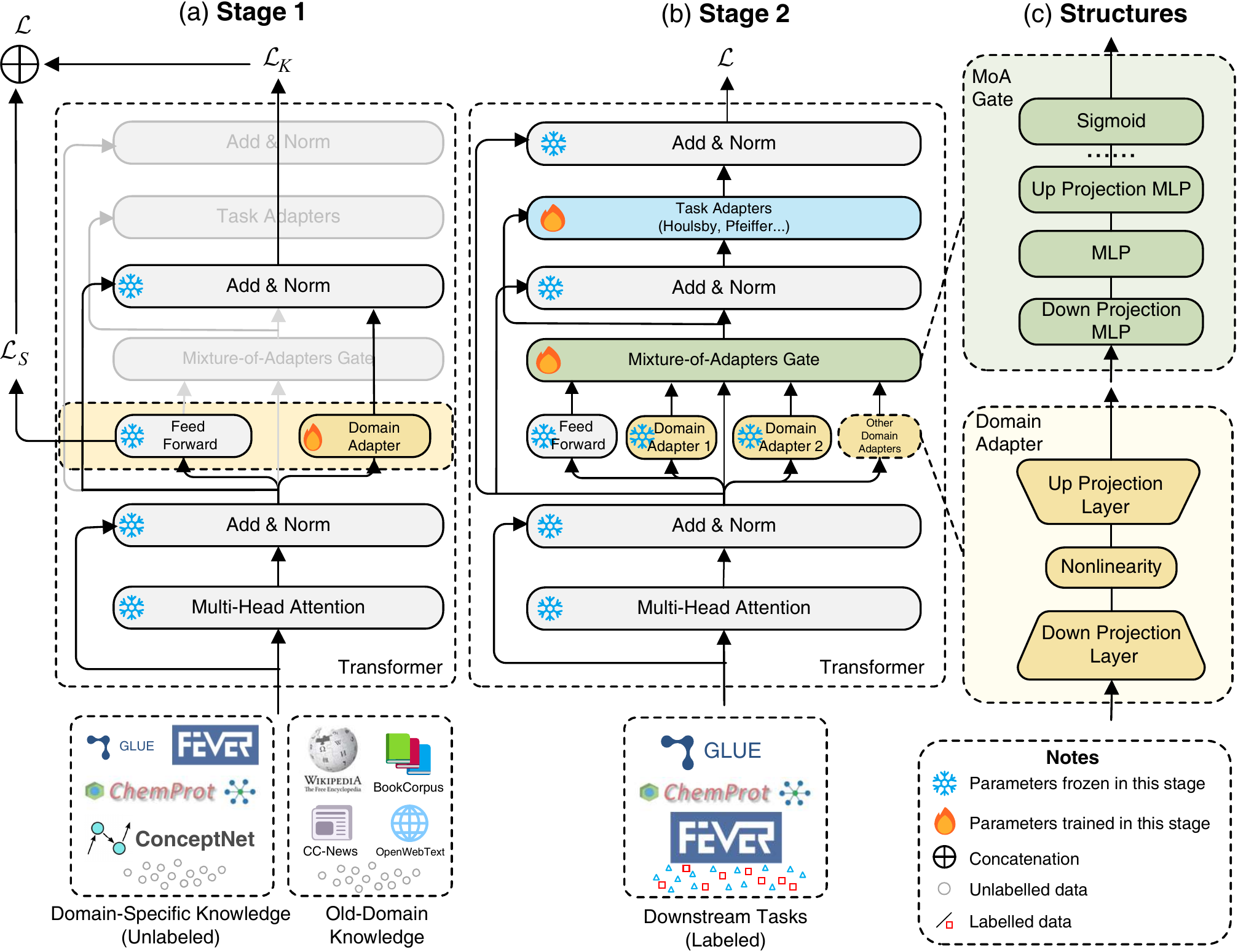}
    \caption{The overall structure of the model. Our training method includes two stages: (a) In Stage 1, we introduce {\ModuleName}s into the model and freeze other parameters. The model learns from domain-specific knowledge (knowledge loss $\mathcal{L}_K$) and keeps similar outputs with the FFN on old-domain knowledge (sample loss $\mathcal{L}_S$). $\mathcal{L}_K$ and $\mathcal{L}_S$ are then combined into the total loss $\mathcal{L}$. (b) In Stage 2, we introduce the mixture-of-adapters gate and task-adapters, then freeze the {\ModuleName}. The model is trained to perform downstream tasks, which gives us the total loss $\mathcal{L}$. (c) shows the detailed structures of the {\ModuleName} and the MoA gate.}
    \label{fig:model}
\vspace{-0.5 em}
\end{figure*}

Given a pre-trained language model $\mathcal{M}$, the input is a sentence $\mathcal{X}=t_1t_2\cdots t_i\cdots t_T$ ($t_i$ indicates the $i$-th token) and the output is the representation of each token. 
The overall architecture of our model is shown in Figure \ref{fig:model}.
The training process is divided into two-stage. 
In Stage 1~(Figure \ref{fig:model} (a)), we inject new feed-forward networks (FFNs) (namely {\ModuleName}) paralleled to the original pre-trained FFNs in some Transformer layers, acting as a key-value memory. 
The newly injected {\ModuleName} is trained on both domain-specific unlabeled data and original pre-training unlabeled data to store new factual associations while keeping old-domain ones. 
All modules are frozen except {\ModuleName} in this stage.
In Stage 2~(Figure \ref{fig:model} (b)), we train a mixture-of-adapters (MoA) gate and a task-adapter on downstream tasks with labeled data, and only these two new modules are updated.
The MoA gate receives outputs from the old-domain FFNs and {\ModuleName}, then outputs a weighted sum of them. 
An additional task-adapter is inserted in each Transformer block to facilitate downstream tasks. 
Figure \ref{fig:model} (c) shows the structures of the {\ModuleName} and the MoA gate.

In this section, we first introduce {\ModuleName}, which learns and stores domain-specific knowledge, and then describe task-adapters that perform the downstream task. 
Finally, we discuss how the MoA gate integrates the outputs from the FFN and the {\ModuleName}.

\subsection{Domain-Adapter}
\label{sub:knowledge-adapter}
Previous studies~\citep{geva2021transformer, decao2021editing, Meng2022Rome} suggest that factual associations are stored in the FFNs of some Transformer layers. 
To help models learn domain-specific knowledge, we propose a lightweight {\ModuleName} that works parallel to the FFNs, and a training method to learn domain-specific knowledge alongside keeping old-domain ones. 
Domain-adapter has a simple bottleneck architecture consisting of a down projection layer, a nonlinearity (such as ReLU~\citep{agarap2018deep}), and an up projection layer. 
This helps keep the parameter size low~\citep{Houlsby2019Feb} with competitive performance.

In Stage 1, the {\ModuleName} is trained with the domain-specific and old-domain datasets in one batch. 
Note that all other parameters are frozen except the {\ModuleName} at this stage. 
Let $\mathcal{L}_K$ denote the \textbf{knowledge loss} related to domain-specific knowledge, and $\mathcal{L}_S$ denote the \textbf{sampling loss} related to old-domain knowledge.
The knowledge loss is a cross-entropy loss on predicting masked tokens, and the sampling loss is designed to align the latent spaces of the old-domain knowledge and new domain-specific knowledge.
The total loss $\mathcal{L}$ is given by a weighted sum of the two, that is:
\begin{equation}
    \mathcal{L} = \lambda\cdot\mathcal{L}_K + \mathcal{L}_S,
\end{equation}
where $\lambda$ is a weight for the knowledge loss.

The \textbf{knowledge loss} is implemented by using cross-entropy loss. 
Given a sentence with $M$ mask tokens whose answers are $m_1,m_2,\cdots, m_M$, respectively, the knowledge loss $\mathcal{L}_K$ is given by
\begin{equation}
    \mathcal{L}_K = -\frac{1}{M} \sum_{i=1}^{M} \log p(m_i),
\end{equation}
where $p(m_i)$ is the probability for token $m_i$ output by $\mathcal{M}$.
Our model accepts two types of domain-specific knowledge as follows, showing improved versatility.
\begin{itemize}[leftmargin=*,label=$\bullet$,noitemsep,partopsep=0pt,topsep=0pt,parsep=0pt]
    \item \textbf{Structured knowledge} If the knowledge dataset is structured (e.g., ConceptNet~\citep{Speer2016ConceptNet}), we translate each relation into a sentence, and then mask out its object. For example, the relation ``the Eiffel tower--/r/LocatedAt--Paris'' is translated into ``The Eiffel Tower is located at Paris.'', then ``Paris'' is substituted with the mask token, and the model is trained to fill the mask.
    \item \textbf{Unstructured knowledge} 
    For unstructured knowledge (e.g., downstream unlabeled texts), we use the masked language model (MLM) similar to RoBERTa pretraining.
    Some tokens are randomly sampled from the input sentence and replaced with the special token \texttt{<mask>}, and the model is trained to predict the masked token.
    The cross-entropy loss is calculated to optimize the model. 
\end{itemize}

For old-domain knowledge and sampling loss, we train the model on general corpora including Wikipedia and BookCorpus~\citep{Zhu2015BookCorpus}. Specifically, for each batch, sentences randomly sampled from the dataset are input into the model. 
Given $L$ layers that have {\ModuleName}s installed, for each such layer $l$, we collect token representations from the FFN $F_l$, and representations from the {\ModuleName} $K_l$. 
The goal is to keep them as similar as possible. 
Thus, we calculate the \textbf{sampling loss} $\mathcal{L}_S$ with L2 loss:
\begin{equation}
    \mathcal{L}_S = \frac{1}{L}\sum_{l=1}^{L} ||F_l-K_l||_2^2.
\end{equation}

\begin{table*}[ht]
\small
\begin{tabularx}{\textwidth}{@{}llXll@{}}
\toprule
\textbf{Domain}  & \textbf{Tasks} & \textbf{Domain Knowledge} & \textbf{\# Tokens} & \textbf{Size} \\ \midrule
Biomed & ChemProt, RCT  & 2.68K papers about biology and chemistry from S2ORC~\citep{lo2020s2orc} & 33.6M & 144MB         \\
Review & Amazon, IMDB & 24.75K randomly selected Amazon reviews & 7.4M  & 34MB          \\
ID & GLUE tasks & Corpus of GLUE tasks & 29.0M  & 146MB         \\
KI & FEVER, CSQA & Corpus of both CommonsenseQA and FEVER datasets & 5.9M  & 34MB          \\ \bottomrule
\end{tabularx}
\vspace{-0.5 em}
\caption{Domain knowledge in Stage 1 training.}
\label{tab:datasets}
\end{table*}

\subsection{Task-Adapter}
After training {\ModuleName}s, the model is aware of the domain knowledge, which is not directly related to downstream tasks though. 
Therefore, we add task-adapters on top of the {\ModuleName} to adapt to downstream tasks. 
For example, a {\ModuleName} trained in biomedical knowledge can support different tasks in the domain, while training it on a task limits its capability to the specific task.
Task-adapters can be any adapter architecture or other parameter-efficient fine-tuning methods, such as the Houlsby adapter~\citep{Houlsby2019Feb}, Pfeiffer adapter~\citep{pfeiffer-etal-2020-mad}, prefix-tuning~\citep{Li2021Prefix}, and so on.
At Stage 2, all parameters other than the task-adapters and the MoA gate (Section \ref{sub:moe}) are frozen. 
The training of the adapter follows its corresponding approach, despite the addition of {\ModuleName}s.
For example, for a text classification task, we add a classification layer on top of the model, freeze all parameters other than the classification layer, the MoA gate, and the task-adapters, feed input texts into the model, and use cross-entropy as the loss.

\subsection{Mixture-of-Adapters Gate}\label{sub:moe}
On downstream tasks, it is possible that the output from the FFN, or a weighted sum of the two, produces better results. 
Therefore, in Stage 2, we train an additional \textbf{mixture-of-adapters} (MoA) gate simultaneously. 
The MoA gate receives the outputs from the attention layer $q$, the {\ModuleName} $K$, and the FFN $F$. $\mathbf{q}$ is first sent into a multi-layer perceptron (MLP):
\begin{equation}
    \mathbf{h} = \mathrm{MLP}(\mathbf{q}).
\end{equation}
The MLP is composed of a down-projection layer $W_d$ and an up-projection layer $W_u$, and $\mathbf{h}=W_u \sigma (W_d \mathbf{q})$, where $\sigma$ is the nonlinearity function.
Then, $\mathbf{h}$ is input into a Sigmoid layer to generate the weights of the FFNs and other {\ModuleName}s:
\begin{equation}
    \mathbf{w}=\mathrm{Sigmoid}(\mathbf{h}).
\end{equation}

The final output $\mathbf{o}$ is a weighted sum of the outputs of the FFNs and the {\ModuleName}:
\begin{equation}
    \mathbf{o} = \mathbf{w}[K;F],
\end{equation}
where $[;]$ denotes matrix concatenation.

\section{Experimental Settings}
In this section, we first introduce the datasets, then the baseline models, the evaluation metrics, and implementation details in the following four subsections, respectively.

\subsection{Datasets}
We conduct experiments on three types of datasets: \textbf{in-domain (ID)} tasks  that require general-domain knowledge; \textbf{out-of-domain (OOD)} tasks that require domain-specific knowledge; \textbf{knowledge-intensive (KI)} tasks that require commonsense knowledge.
\begin{itemize}[leftmargin=*,label=$\bullet$,noitemsep,partopsep=0pt,topsep=0pt,parsep=0pt]
\item \textbf{ID}: GLUE Benchmark~\citep{wang2018glue} including MNLI~\citep{williams2017broad}, CoLA~\citep{warstadt2019neural}, MRPC~\citep{dolan2005automatically}, SST-2~\citep{socher2013recursive}, RTE~\citep{dagan2005pascal,haim2006second,giampiccolo2007third,bentivogli2009fifth}, STS-B~\citep{cer2017semeval}, WNLI~\citep{levesque2012winograd}, QNLI~\citep{rajpurkar2016squad}, and QQP~\citep{iyer2017first}. 
\item \textbf{OOD}: ChemProt~\citep{Kringelum2016Jan}, RCT~\citep{lee2017pubmed}, IMDB~\citep{maas2011learning}, and Amazon~\citep{he2016ups}. 
ChemProt is a manually annotated chemical-protein interaction dataset extracted from 5,031 abstractions. 
RCT is a dataset based on PubMed for sentence classification. 
IMDB provides 25,000 movie reviews for sentiment analysis. 
Amazon is a dataset containing product reviews from Amazon, annotated with user ratings.
\item \textbf{KI}: FEVER~\citep{Thorne18Fever} and CommonsenseQA~(CSQA)~\citep{talmor2019commonsenseqa}. 
FEVER consists of 185,445 claims that correspond to Wikipedia articles and are classified as supported, refuted, and not enough information. 
CommonsenseQA consists of 12,247 questions with 5 choices and requires commonsense knowledge to predict the correct answers.
\end{itemize}

For Stage 1, we train the {\ModuleName} with unstructured knowledge related to the dataset following Section \ref{sub:knowledge-adapter}. 
The unstructured knowledge used is listed in Table \ref{tab:datasets}.
We also experiment with structured knowledge in Section~\ref{sub:structured}.
For Stage 2, we adopt the true few-shot setting following~\citep{perez2021true} to demonstrate the effectiveness of {\ModelNameShort}.
For each dataset class, we randomly sample $K=16$ examples from the original training set as the new training set, and another different $K=16$ examples as the validation set.
The original validation set will be used as the test set.
The Pfeiffer adapter is used in Stage 2 unless stated otherwise.

\begin{table*}[ht]
\centering
\small
\begin{tabularx}{\textwidth}{@{}cc*{6}{>{\centering\arraybackslash}X}@{}}
\toprule
\multicolumn{1}{l}{}            & \textbf{Datasets} & \textbf{HO}                & \textbf{PF}                        & \textbf{LO}                & \textbf{PT}               & \textbf{FT}                        & \textbf{\ModelNameShort}                        \\ \midrule
\multirow{5}{*}{OOD}    & ChemProt         & 47.1\tiny{±12.2} & 53.7\tiny{±8.2}                       & 24.2\tiny{±11.6} & 17.1\tiny{±7.3} & 57.9\tiny{±4.0}          & \textbf{60.6\tiny{±4.9}} \\
                                & RCT               & 25.2\tiny{±2.6}  & 21.9\tiny{±2.5}          & 18.5\tiny{±3.7}  & 24.4\tiny{±3.7} & 21.0\tiny{±3.2}          & \textbf{26.4\tiny{±0.8}} \\
                                & IMDB              & 56.0\tiny{±5.7}  & 55.4\tiny{±5.6}          & 43.7\tiny{±7.7}  & 53.3\tiny{±2.6} & 46.3\tiny{±13.7}         & \textbf{58.1\tiny{±5.1}} \\
                                & Amazon            & 48.8\tiny{±3.2}  & 49.7\tiny{±1.4}          & 51.5\tiny{±3.9}  & 52.7\tiny{±2.8} & 51.7\tiny{±6.2}          & \textbf{54.7\tiny{±1.6}} \\
                                & \textbf{Avg.}     & 44.3                      & 45.2                              & 34.5                      & 36.9                     & 44.2                              & \textbf{50.0}                     \\ \midrule
\multirow{10}{*}{ID}  & MNLI              & 37.2\tiny{±0.3}  & 35.7\tiny{±0.1}          & 34.8\tiny{±1.5}  & 35.4\tiny{±0.0} & 35.3\tiny{±0.2}          & \textbf{37.3\tiny{±1.5}} \\
                                & COLA              & 17.6\tiny{±5.5}  & 9.1\tiny{±5.0}           & 7.1\tiny{±3.0}   & 12.1\tiny{±5.5} & \textbf{21.3\tiny{±1.7}} & 20.1\tiny{±6.3}          \\
                                & MRPC              & 81.2\tiny{±0.2}  & 80.7\tiny{±0.6}          & 64.0\tiny{±20.5} & \textbf{81.6\tiny{±0.5}} & 81.3\tiny{±0.1}          & \textbf{81.6\tiny{±0.5}} \\
                                & SST2              & 54.7\tiny{±3.6}  & 53.3\tiny{±1.9}          & 50.5\tiny{±1.0}  & 52.5\tiny{±1.2} & 54.8\tiny{±1.4}          & \textbf{56.4\tiny{±3.5}} \\
                                & RTE               & 53.5\tiny{±1.4}  & 54.1\tiny{±1.3}          & 53.4\tiny{±2.1}  & 53.4\tiny{±1.1} & 54.7\tiny{±1.3}          & \textbf{54.9\tiny{±1.5}} \\
                                & STS-B              & 88.1\tiny{±1.6}  & \textbf{90.6\tiny{±0.1}} & 89.5\tiny{±0.8}  & 85.6\tiny{±4.1} & 78.4\tiny{±8.0}          & 89.8\tiny{±0.4}          \\
                                & WNLI              & 57.3\tiny{±1.3}  & 58.1\tiny{±2.5}          & 59.1\tiny{±1.2}  & 57.3\tiny{±0.7} & 58.7\tiny{±1.8}          & \textbf{60.1\tiny{±1.8}} \\
                                & QNLI              & 53.0\tiny{±0.1}  & 51.9\tiny{±0.8}          & 53.3\tiny{±1.3}  & 52.1\tiny{±0.9} & 51.3\tiny{±0.1}          & \textbf{54.8\tiny{±1.8}} \\
                                & QQP               & 54.7\tiny{±0.6}  & 55.2\tiny{±1.0}          & 53.0\tiny{±2.0}  & 55.3\tiny{±0.3} & 55.3\tiny{±0.3}          & \textbf{56.1\tiny{±0.6}} \\
                                & \textbf{Avg.}     & 55.3                      & 54.3                              & 51.6                      & 53.9                     & 54.6                              & \textbf{56.8}                     \\ \midrule
\multirow{3}{*}{KI} & FEVER             & 20.2\tiny{±4.3}  & 27.4\tiny{±7.5}          & 22.6\tiny{±10.6} & 31.1\tiny{±3.6} & \textbf{36.1\tiny{±6.7}}          & 32.6\tiny{±9.4} \\
                                & CSQA     & 27.3\tiny{±0.7}  & 34.1\tiny{±8.7}          & 20.3\tiny{±10.9} & 29.6\tiny{±4.2} & 29.6\tiny{±3.0}          & \textbf{38.9\tiny{±4.0}} \\
                                & \textbf{Avg.}     & 23.8                      & 30.8                              & 21.5                      & 30.4                     & 32.9                              & \textbf{35.8}                     \\ \midrule
\multicolumn{1}{l}{}            & \textbf{Avg.}     & 48.1                      & 48.7                              & 43.0                      & 46.2                     & 48.9                              & \textbf{52.2}                     \\ \bottomrule
\end{tabularx}
\caption{The overall performance of single {\ModelNameShort} and baselines on the downstream tasks. We use $K=16$ (per class) for few-shot experiments. The best result for each dataset is made bold. We report mean and standard deviation over 3 runs with different random seeds.}
\label{tab:single-perf}
\end{table*}

\subsection{Baselines}

In our experiments, we use the following models as the main baselines. 
For convenience, we refer to them with the abbreviations in the parentheses later.
\begin{itemize}[leftmargin=*,label=$\bullet$,noitemsep,partopsep=0pt,topsep=0pt,parsep=0pt]
    \item \textbf{\textsc{Houlsby} (HO)}: Houlsby adapter~\citep{Houlsby2019Feb} plugged into the RoBERTa-large model for downstream tasks. Only adapter parameters are trained. It adds two adapter blocks consisting of bottleneck networks in each Transformer block.

    \item \textbf{\textsc{Pfeiffer} (PF)}: Pfeiffer adapter~\citep{pfeiffer-etal-2020-mad} plugged into the RoBERTa-large model. 
    This is similar to the Houlsby adapter, but with a different architecture. 
    Pfeiffer adapter has only one adapter layer in each Transformer block, while Houlsby has two. 
    Also, Pfeiffer makes minor tweaks in the adapter architecture, such as the layer norm and nonlinearity.

    \item \textbf{\textsc{LoRA} (LO)}: LoRA~\citep{Hu2021LoRA} applied to the RoBERTa-large model. 
    LoRA freezes the MLP modules and represents updates to the attention weights with two low-rank matrices, thus saving space.

    \item \textbf{\textsc{Prefix-Tuning} (PT)}: Prefix-Tuning~\citep{Li2021Prefix} with the RoBERTa-large model. 
    Prefix-Tuning trains a number of prompt embeddings for each task and pre-pends it before tokens.

    \item \textbf{\textsc{Fine-Tuning} (FT)}: Fine-tuning all of the parameters of the RoBERTa-large model on downstream tasks.
\end{itemize}

\subsection{Evaluation Metrics}
We adopt the Pearson correlation for STS-B since it is a regression task. 
The remaining are text classification tasks. 
Following~\citet{wang2018glue, gururangan2020don, diao2021taming}, we adopt macro-F1 for MRPC and QQP, and micro-F1 for others as evaluation metrics. 
Macro-F1 computes the F1 independently for each metric, while micro-F1 computes an average metric of all classes.
To account for the instability of small datasets, we report the average performance and the standard deviation of 3 runs with different random seeds.

\begin{table*}[t]
\centering
\small
\begin{tabularx}{\textwidth}{@{}l*{8}{>{\centering\arraybackslash}X}@{}}
\toprule
                  & Amazon                                 & IMDB                               & FEVER                              & WNLI                               & QQP                                & RTE                                & MRPC                               & \textbf{Avg.}  \\ \midrule
\textbf{Pfeiffer} & 49.7\tiny{±1.4}          & 55.4\tiny{±5.6}          & 27.4\tiny{±7.5}          & 58.1\tiny{±2.5}          & 55.2\tiny{±1.0}          & 54.1\tiny{±1.3}          & 80.7\tiny{±0.6}          & 54.4          \\
\textbf{Single}   & \textbf{54.7\tiny{±1.6}} & \textbf{58.1\tiny{±5.1}} & 32.6\tiny{±9.4} & \textbf{60.1\tiny{±1.7}} & 56.1\tiny{±0.6} & \textbf{54.9\tiny{±1.5}} & \textbf{81.6\tiny{±0.5}} & \textbf{56.9} \\
\textbf{Parallel} & 51.6\tiny{±2.4}          & 47.9\tiny{±2.1}          & \textbf{34.5\tiny{±0.5}}          & 58.7\tiny{±1.8}          & \textbf{57.8\tiny{±3.5}}          & 53.8\tiny{±0.9}          & 81.0\tiny{±0.2}          & 55.0          \\ \bottomrule
\end{tabularx}
\caption{The performance of parallel {\ModuleName}s on the chosen downstream tasks. Parallel, Single, and Pfeiffer denote parallel {\ModuleName}s, single {\ModuleName}, and vanilla RoBERTa + Pfeiffer, respectively. The best result for each dataset is made bold.}
\vspace{-0.5 em}
\label{tab:parallel}
\end{table*}

\subsection{Implementation}
We implement our RoBERTa-large model based on the Transformers library from HuggingFace\footnote{\url{https://github.com/huggingface/transformers}}. 
The Houlsby adapter, the Pfeiffer adapter, and Prefix-Tuning are implemented based on the adapter-transformers library~\citep{pfeiffer2020AdapterHub}. 
LoRA is implemented based on OpenDelta~\citep{ding2022delta}.
During Stage 1, we train the {\ModuleName} with learning rate 1e-4, batch size 20, and weight decay 0.05. 
The knowledge loss factor $\lambda$ is set to 0.5. 
We train the 7 and 11 layers of RoBERTa-large with {\ModuleName} in 10 epochs. 
In Stage 2, we use the Pfeiffer adapter as the default task-adapter and train 20 epochs.
All the experiments are conducted on Nvidia 2080Ti GPUs.
We find the best hyper-parameters through grid search and the best results are listed in Appendix \ref{sec:setup}.
The computation time can be found in Appendix \ref{sec:budget}.

\begin{table*}[]
\small
\centering
\begin{tabularx}{\textwidth}{@{}l*{7}{>{\centering\arraybackslash}X}@{}}
\toprule
\textbf{Datasets}     & ChemProt                                 & IMDB                               & MRPC                               & STS-B                               & CSQA                               & \textbf{Avg.}  \\ 
\midrule
\textbf{\ModelNameShort}           & \textbf{60.6\tiny{±4.9}} & \textbf{58.1\tiny{±5.1}} & \textbf{81.6\tiny{±0.5}} & 89.8\tiny{±0.4} & \textbf{38.8\tiny{±4.0}} & \textbf{65.8} \\ 
\midrule
\textbf{-- MoA}       & 55.7\tiny{±1.8}          & 49.8\tiny{±0.0}          & 80.9\tiny{±0.5}          & 88.4\tiny{±0.4}          & 28.3\tiny{±1.3}          & 60.6          \\
\textbf{-- Old}       & 54.3\tiny{±6.5}          & 41.4\tiny{±3.6}          & 78.7\tiny{±3.7}          & \textbf{90.0\tiny{±0.4}}          & 27.1\tiny{±0.3}          & 58.3          \\
\textbf{-- DA} & 21.2\tiny{±4.7}          & 56.8\tiny{±3.9}          & 81.0\tiny{±0.5}          & 80.8\tiny{±2.4}          & 27.4\tiny{±0.5}          & 53.4          \\
\midrule
\textbf{AdapterFusion}           & 47.7\tiny{±0.1} & 54.4\tiny{±2.0} & 78.0\tiny{±1.5} & 90.3\tiny{±0.3} & 25.0\tiny{±1.7} & 59.1 \\ 
\textbf{K-Adapter}           & 58.2\tiny{±5.0} & 55.6\tiny{±4.5} & 53.9\tiny{±5.9} & 89.7\tiny{±0.4} & 26.2\tiny{±4.7} & 56.7 \\ 
\textbf{CPT}           & 45.9\tiny{±0.3} & 56.1\tiny{±5.2} & 81.0\tiny{±0.5} & 90.2\tiny{±0.1} & 33.7\tiny{±2.7} & 61.4 \\ 
\bottomrule
\end{tabularx}
\caption{Ablations of the MoA gate, old-domain knowledge, and the {\ModuleName} structure and comparisons with other adapter-based tuning methods.
For \textbf{-- Old}, we omit old-domain knowledge in Stage 1 training.
For \textbf{-- DA}, we remove the {\ModuleName} structure and conduct both stages of training only with Pfeiffer adapters.
The best results for each dataset are made bold.
}
\label{tab:ablation}
\end{table*}

\begin{table}[t]
\scriptsize
\centering
\begin{tabularx}{\columnwidth}{@{}lccccc@{}}
\toprule
\textbf{Datasets}         & MRPC                               & STS-B           & FEVER                                         & CSQA                               & \textbf{Avg.}  \\ \midrule
\textbf{\ModelNameShort}               & 81.6\tiny{±0.5}          & 89.8\tiny{±0.4}          & 20.2\tiny{±4.3}                     & 38.8\tiny{±4.0}          & 57.6          \\
\textbf{+ ConceptNet} &  \textbf{81.7\tiny{±0.3}} & \textbf{90.1\tiny{±0.1}} & \textbf{30.5\tiny{±3.1}} & \textbf{40.0\tiny{±0.2}} & \textbf{60.6} \\ \bottomrule
\end{tabularx}
\caption{The results of {\ModelNameShort} trained on structured and unstructured knowledge. \textbf{+ ConceptNet} stands for {\ModuleName}s trained on both the unstructured knowledge and ConceptNet.}
\vspace{-2 em}
\label{tab:unstructured}
\end{table}

\begin{table*}[ht]
\small
\centering
\begin{tabularx}{\textwidth}{@{}l*{6}{>{\centering\arraybackslash}X}@{}}
\toprule
\textbf{Datasets} & ChemProt                                 & IMDB                               & MRPC                               & STS-B                               & CSQA                               & \textbf{Avg.}  \\ \midrule
Houlsby           & 47.1\tiny{±12.2}         & 48.1\tiny{±4.5}          & 80.0\tiny{±1.5}          & 86.6\tiny{±3.2}          & 35.8\tiny{±8.9}          & 59.5          \\
Prefix-Tuning     & 17.1\tiny{±7.3}          & 39.1\tiny{±7.2}          & 81.6\tiny{±0.4}          & 88.6\tiny{±0.5}          & 33.3\tiny{±0.0}          & 51.9          \\
LoRA              & 19.5\tiny{±11.1}         & 36.1\tiny{±4.1}          & 81.2\tiny{±0.0}          & 86.7\tiny{±1.4}          & 20.3\tiny{±10.9}         & 48.8          \\
Pfeiffer          & \textbf{60.6\tiny{±4.9}} & \textbf{58.1\tiny{±5.1}} & \textbf{81.6\tiny{±0.5}} & \textbf{89.8\tiny{±0.4}} & \textbf{38.8\tiny{±4.0}} & \textbf{65.8} \\ \bottomrule

\end{tabularx}
\caption{The results of {\ModelNameShort} combined with different kinds of task-adapters. By default, we use Pfeiffer in previous experiments.}
\label{tab:adapters}
\vspace{-2 em}
\end{table*}

\section{Experimental Results}
\label{sec:expr}
We compare the performance of {\ModelNameShort} with our baselines on 15 datasets. 
First, we train the {\ModuleName} for each domain individually and then perform each task with its corresponding {\ModuleName}, which shows significant improvement over our baselines. 
Next, we plug in several {\ModuleName}s trained on different domains parallelly to verify the scalability of our model.

\subsection{Single Domain Adapter}

Table~\ref{tab:single-perf} shows the performance of a single domain adapter compared with baselines. 
It is only trained on unstructured knowledge during Stage 1 in the following experiments. 
Results show that {\ModelName} outperforms our baselines in most datasets, with an average of 3.5\% improvement over the best baseline adapter (i.e., Pfeiffer), and 3.3\% over fine-tuning.
Our method even outperforms fine-tuning in most datasets, despite far less training time and smaller parameter size. 
Over the datasets, {\ModelNameShort} shows the most significant improvement on ChemProt, with 6.9\% over Pfeiffer and 2.7\% over fine-tuning. 
One possible reason is that {\ModelNameShort} learns the necessary knowledge to detect the chemical-protein interaction. 
For example, {\ModelNameShort} shows more familiarity with words associated with that field, such as ``gefitinib'' and ``tyrosine kinase inhibitor''. 
In contrast, {\ModelNameShort} falters on STS-B, falling behind Pfeiffer by 0.8\%. This is because the knowledge in Stage 1 is not effectively utilized. STS-B consists of sentence pairs like ``The cat sat on the mat'' and ``The cat did not sit on the mat'', with little need for additional knowledge.
Across the three task domains, {\ModelNameShort} has an average improvement of 4.8\% over RoBERTa + Pfeiffer on out-of-domain tasks, 2.5\% on in-domain tasks, and 5.0\% on knowledge-intensive tasks. 
It shows that {\ModelNameShort} is not only effective for out-of-domain tasks and knowledge-intensive tasks that require additional knowledge but is helpful for general-domain language tasks as well, demonstrating its ability to excel at both in-domain and out-of-domain tasks (reliability).

\subsection{Parallel Domain Adapters}
In the previous section, we explored using a single {\ModuleName} for each downstream task.
Next, we show the scalability of {\ModelNameShort} by using parallel {\ModuleName}s and only train the MoA layer and task-adapters in Stage 2.
The training process in Stage 2 follows the previous experiments.
Table \ref{tab:parallel} shows the comparison across single {\ModuleName}, parallel {\ModuleName}s, and RoBERTa + Pfeiffer on 7 datasets.
On average, parallel {\ModuleName}s show an improvement of 0.6\% over vanilla RoBERTa + Pfeiffer, even though they fall behind the single domain adapter by 1.9\%. 
This could be attributed to the MoA gate choosing the sub-optimal {\ModuleName} for some test data.
Still, considering its improvement over Pfeiffer, the MoA gate chooses the correct {\ModuleName} in most cases.
Therefore, {\ModelNameShort} demonstrates its scalability, allowing end users to train Stage 1 on different datasets and combine them later.
Overall, in both single and parallel situations, {\ModelNameShort} significantly improves upon the vanilla RoBERTa + Pfeiffer model with a small increase in model size.
This is due to the ability of {\ModelNameShort} to capture knowledge and the MoA to select useful knowledge for downstream tasks.

\section{Analysis}
In this section, we analyze the respective contributions of each part of {\ModelNameShort} through detailed analysis, including the Stage 1 training, task-adapters in Stage 2, and the mixture-of-adapters gate.

\subsection{Ablation Study}
In this section, we conduct an ablation study to reveal the contributions of each part of the model.
There are three variants:
(1)
We remove the MoA gate and choose the {\ModuleName} instead of the RoBERTa feed-forward layer (--MoA).
(2)
We exclude old-domain knowledge during Stage 1 (--Old).
(3)
To examine whether the training procedures, rather than the {\ModelNameShort} structure, contribute the most to our results,
we conduct Stage 1 and Stage 2 training only with task-adapters (--DA).
Table~\ref{tab:ablation} shows the results of the ablation study.
As expected, the average performance drops in all three settings. 
Without MoA gate, old-domain knowledge FFNs, and structure knowledge, it is observed a drop of 5.2\%, 7.5\%, and 12.4\%, respectively, showing that the MoA gate, the old-domain knowledge, and the {\ModelNameShort} structure are all fundamental in the model. 
Relatively, the MoA has the smallest impact because the old-domain knowledge in Stage 1 can also help the model retain the knowledge in RoBERTa. 
The {\ModuleName} has the largest impact since it only stores domain knowledge and can keep it during Stage 2. 
In contrast, conducting Stage 1 and 2 training on the Pfeiffer adapter causes catastrophic forgetting.

\subsection{Structured and Unstructured Knowledge}\label{sub:structured}
In Section \ref{sec:expr}, the {\ModelNameShort} is only trained on unstructured knowledge. 
As a comparison, we also train the domain adapter on ConceptNet, a structured knowledge dataset, and then attach both the unstructured and structured to our model and train the MoA layer and the task-adapter during Stage 2.

Table~\ref{tab:unstructured} shows the result of combining structured and unstructured knowledge in Stage 1. 
FEVER and CSQA, two knowledge-intensive tasks, have the greatest improvement: 10.3\% for FEVER and 1.2\% for CSQA. This is because ConceptNet stores commonsense knowledge that can help both tasks. 
Meanwhile, MRPC and STS-B also obtain improvement, showing that ConceptNet can benefit general language tasks as well. 
In conclusion, the experiment demonstrates the ability of {\ModelNameShort} to utilize structured knowledge, the extensibility of our model, and the possible benefits of structured knowledge.

\subsection{Effectiveness of Task-Adapters}

In most experiments of this paper, we adopt Pfeiffer as the task-adapter unless otherwise specified.
In this section, we test the performance of {\ModelNameShort} combined with other kinds of task-adapters, including Houlsby, Prefix-Tuning, LoRA, and Pfeiffer.
Table~\ref{tab:adapters} gives the result of different task-adapters. 
Pfeiffer surpasses others by at least 6.3\%. 
Even though Houlsby is on par with Pfeiffer, Pfeiffer only requires half the number of newly introduced parameters compared to Houlsby, making it the optimal choice of task-adapters in our experiment.

\section{Conclusion}
In this paper, we proposed {\ModelNameShort}, a mixture of adapters for domain adaptation.
We first decouple the knowledge modules (i.e., FFNs) into the old-domain and domain-specific FFNs.
Then we propose a two-stage adapter tuning strategy: first tuning the domain adapter on each domain and then tuning the task adapter on each task.
Moreover, our model could be scaled to multiple domains easily with the introduction of the mixture-of-adapters gate.
Empirically, {\ModelNameShort} achieved significant improvement over in-domain tasks, out-of-domain tasks, and knowledge-intensive tasks.
Further analyses demonstrate the reliability, scalability, and efficiency of our method.

\section*{Limitations}
Although {\ModelNameShort} achieves promising results on domain adaptation compared with baseline models, there are certain limitations.
{\ModelNameShort} is a two-stage approach, which is not fully end-to-end.
Our approach requires training a domain adapter and task adapter, respectively.
In the future, we will explore the unifying domain and task adapters by merging them into one.

\section*{Acknowledgments}
We thank the anonymous reviewers for their valuable suggestions.
This work was supported by the General Research Fund (GRF) of Hong Kong (No. 16310222 and No. 16201320). 
Shizhe Diao and Ruijia Xu were supported by the Hong Kong Ph.D. Fellowship Scheme (HKPFS).

\bibliography{custom}
\bibliographystyle{acl_natbib}

\clearpage

\appendix
\section{Experimental Setup}\label{sec:setup}
Our domain adapter has a reduction factor of $16$, consisting of two linear layers $4096 \times 256$ and $256 \times 1024$ (1.31M parameters). With each domain adapter also comes a MoA gate which has an FFN with $4096\times 2$ (number of {\ModelNameShort}s) parameters. Since domain adapters are placed in Layers 7 and 11, they have 2.6M parameters in total. Therefore, the domain adapters (excluding task-adapters) only add 0.7\% additional parameters to RoBERTa-large.

We preprocess the unstructured data in Stage 1 similar to the masked language model directive in RoBERTa. From the text, we choose 15\% of tokens uniformly to perform possible alterations. In those tokens, 85\% are replaced with \texttt{<mask>}, 10\% are left unchanged, and 5\% are replaced with a random token. The preprocessing is implemented with \texttt{DataCollatorForLanguageModeling} in Huggingface Transformers. In Stage 2, we use few-shot setting with $K=16$. For each class of the dataset, we randomly select $16$ examples before run.

In Stages 1 and 2, we use a linear weight scheduler. All the models are optimized by AdamW~\citep{loshchilov2017adamw} with weight decay 0.05. The best hyperparameters for Stage 2 are found with grid search, with batch size $\{2, 4, 8, 16\}$ and learning rate $\{5e-5, 1e-4, 5e-4\}$. The details can be found in Tables \ref{tab:stage1_setup} and \ref{tab:stage2_setup}.

\section{Computational Budget}\label{sec:budget}
Stage 1 training takes relatively longer time, while Stage 2 is fast due to the few-shot setting. The training time of Stage 1 is proportional to the number of tokens. For reference, with 4 Nvidia RTX 2080Ti, Stage 1 training for Biomed (33.6M tokens) takes \textasciitilde 45min per epoch, and training for Review (7.4M tokens) takes \textasciitilde 5min per epoch. Stage 2 training is generally fast: The 20-epoch training process usually takes less than 5min with 4 Nvidia RTX 2080Ti.

\section{Details of Datasets}
We conduct experiments on three types of datasets: \textbf{in-domain (ID)} tasks  that require general-domain knowledge; \textbf{out-of-domain (OOD)} tasks that require domain-specific knowledge; \textbf{knowledge-intensive (KI)} tasks that require commonsense knowledge.

For in-domain tasks, we evaluate our model on the GLUE Benchmark~\citep{wang2018glue}. 
It includes MNLI~\citep{williams2017broad}, CoLA~\citep{warstadt2019neural}, MRPC~\citep{dolan2005automatically}, SST-2~\citep{socher2013recursive}, RTE~\citep{dagan2005pascal,haim2006second,giampiccolo2007third,bentivogli2009fifth}, STS-B~\citep{cer2017semeval}, WNLI~\citep{levesque2012winograd}, QNLI~\citep{rajpurkar2016squad}, and QQP~\citep{iyer2017first}. 
They are all single-sentence or sentence pair classification tasks except STS-B, which is a regression task. 

We also evaluate our model on several out-of-domain tasks, including ChemProt~\citep{Kringelum2016Jan}, RCT~\citep{lee2017pubmed}, IMDB~\citep{maas2011learning}, and Amazon~\citep{he2016ups}. 
ChemProt is a manually annotated chemical-protein interaction dataset extracted from 5,031 abstractions. 
RCT is a dataset based on PubMed for sentence classification. 
IMDB provides 25,000 movie reviews for sentiment analysis. 
Amazon is a dataset containing product reviews from Amazon, annotated with user ratings.

For knowledge-intensive tasks, we evaluate our model on \textbf{FEVER}~\citep{Thorne18Fever} and \textbf{CommonsenseQA} (CSQA)~\citep{talmor2019commonsenseqa}. 
FEVER consists of 185,445 claims that correspond to Wikipedia articles and are classified as supported, refuted, and not enough information. 
CommonsenseQA consists of 12,247 questions with 5 choices, each of which requires commonsense knowledge to predict the correct answers.

For Stage 1, we train {\ModuleName}s with unstructured knowledge related to the dataset following Section \ref{sub:knowledge-adapter}. 
The unstructured knowledge used is listed in Table \ref{tab:datasets}.
We also experiment with structured knowledge in Section~\ref{sub:structured}.
For Stage 2, we adopt the true few-shot setting following~\citep{perez2021true} to demonstrate the effectiveness of {\ModelNameShort}.
For each class of each dataset, we randomly sample $K=16$ examples from the original training set as the new training set, and another different $K=16$ examples as the validation set.
The original validation set will be used as the test set.
The Pfeiffer adapter is used in Stage 2 unless stated otherwise.

\begin{table*}[ht]
\centering
\begin{tabular}{c|c}
\toprule
Config & Value \\
\hline
Optimizer & AdamW \\
Learning rate & 1e-4 \\
Weight decay & 0.05 \\
Optimizer momentum & $\beta_1, \beta_2{=}0.9, 0.999$ \\
Batch size & \{2, 5\} \\
Learning rate schedule & linear decay \\
training epochs & 10 \\
\bottomrule
\end{tabular}
\caption{Stage 1 training: experimental setup.}
\label{tab:stage1_setup} 
\end{table*}

\begin{table*}[ht]
\centering
\begin{tabular}{c|c}
\toprule
Config & Value \\
\hline
Optimizer & AdamW \\
Learning rate & \{5e-5, 1e-4, 5e-4\} \\
Weight decay & 0.05 \\
Optimizer momentum & $\beta_1, \beta_2{=}0.9, 0.999$ \\
Batch size & \{2, 4, 8, 16\} \\
Learning rate schedule & linear decay \\
warmup epochs & 2 \\
training epochs & 20 \\
\bottomrule
\end{tabular}
\caption{Stage 2 training: experimental setup.}
\label{tab:stage2_setup} 
\end{table*}

\end{document}